\documentclass{sig-alternate-05-2015}

\usepackage{graphicx}        
\usepackage{epsfig}
\usepackage{multirow}
\usepackage{booktabs, caption,fixltx2e}
\usepackage[flushleft]{threeparttable}
\usepackage{amsmath}
\usepackage{amssymb}
\usepackage{wrapfig}
\usepackage{lmodern}

\makeatletter
\def\hlinew#1{%
	\noalign{\ifnum0=`}\fi\hrule \@height #1 \futurelet
	\reserved@a\@xhline}

\def\sharedaffiliation{%
\end{tabular}
\begin{tabular}{c}}

\begin{document}


\doi{}

\isbn{1}


%
\conferenceinfo{}{}

\title{How smart does your profile image look? Estimating intelligence from social network profile images}
%

%
%
%
%
%

\numberofauthors{2} 
%
\author{
%
%
\alignauthor Xingjie Wei$^{1,2}$\\
       \email{x.wei@bath.ac.uk}\\
\alignauthor David Stillwell$^{1}$\\
       \email{ds617@cam.ac.uk}\\
        \sharedaffiliation
        \affaddr{Psychometrics Centre, University of Cambridge, Cambridge, CB2 1AG, U.K.$^{1}$}\\
        \affaddr{School of Management, University of Bath, Bath, BA2 7AY, U.K.$^{2}$}
}


\clubpenalty=10000 
\widowpenalty = 10000

\maketitle


\begin{abstract}
Profile images on social networks are users' opportunity to present themselves and to affect how others judge them. We examine what Facebook images say about users' perceived and measured intelligence. 1,122 Facebook users completed a matrices intelligence test and shared their current Facebook profile image. Strangers also rated the images for perceived intelligence. We use automatically extracted image features to predict both measured and perceived intelligence. Intelligence estimation from images is a difficult task even for humans, but experimental results show that human accuracy can be equalled using computing methods. We report the image features that predict both measured and perceived intelligence, and highlight misleading features such as ``smiling'' and ``wearing glasses'' that are correlated with perceived but not measured intelligence. Our results give insights into inaccurate stereotyping from profile images and also have implications for privacy, especially since in most social networks profile images are public by default.
\end{abstract}

%
%
%


\keywords{Intelligence quotient; Measured intelligence; Perceived intelligence; Intelligence estimation; Computational aesthetics; IQ}

\section{Introduction}
\label{sec:intro}
Profile images are fundamental to online social networks. They are typically displayed each time a user posts a message or shares a piece of content, and are normally public by default. As such, they are an important avenue for users to share their self-representation, and they can have a big effect on how friends and strangers judge the user. Recent experimental research \cite{baert2015do} shows that Facebook profile images affect employers' hiring decisions: Candidates with the most attractive profile images obtained 39\% more job interview invitations than those with the least attractive images.

General profile images include behavioural cues such as camera views, poses, clothes, locations and the presence of friends and partners; or preference cues such as pets, cartoon characters, sports equipment and special signs, which are linked to people's lifestyles and how they represent themselves in the social network. Such cues are the outcome of intentional choices, driven by psychological differences. 

Intelligence is an individual difference that is related to important life outcomes such as income \cite{zagorsky2007do} and relationships \cite{keller2013the}. The first impressions of intelligence can have significant consequences in social scenarios such as employment. Different from other \textit{neutral} psychological traits such as personality, where there is no ideal personality, high intelligence is a trait that people want to project to others by self-representation. 

In this paper, we investigate whether a user's intelligence can be inferred from profile images. This will help us to better understand human behaviour, help Web users to better manage their self-representation, and enable an understanding of the privacy risk of having a public profile image. There are very few works studying the relationship between images of people and their intelligence \cite{kleisner2014perceived,talamas2016blinded,talamas2016eyelid}, and those that exist focus on facial photographs using small sample sizes. Our work is distinguished from those by taking into account the general profile images which contain more psychological and behavioural cues, such as the image background, and using a much larger data set.

We define a user's intelligence score measured by an IQ test as \textit{measured intelligence} (MI) and the intelligence score rated by human observers' perceptions based on the self-representation of users as \textit{perceived intelligence} (PI). We propose a framework of intelligence estimation from profile images by both humans and computers as shown in Figure \ref{fig-fw}. We use profile images from 1,122 Facebook users who have taken an IQ test, and whose profile images have also been rated by strangers. We extract visual features from profile images and examine the relationships between those features and both MI and PI. Our contributions can be summarised as follows:

\begin{figure*}[tb]
	\centerline{\epsfig{figure=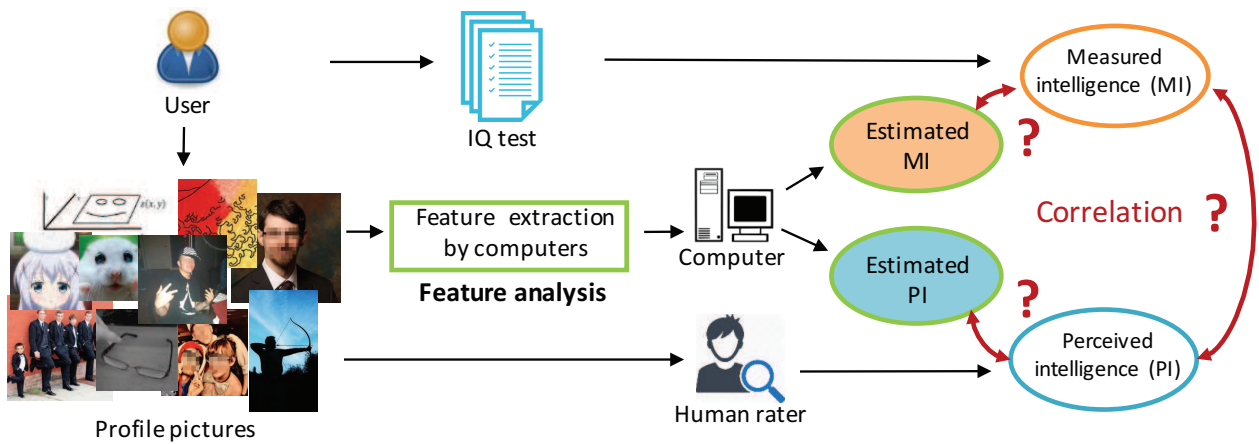, width=0.95\textwidth}}
	\caption{Illustration of image based intelligence estimation.}
	\label{fig-fw}
\end{figure*}

\begin{itemize}
	\item We designed experiments to investigate how humans make intelligence judgments (i.e., PI) for others from profile images. We found different people have relatively consistent judgements and such judgements are associated with MI for both men and women.
	\item We proposed a system of automatic intelligence estimation from profile images and evaluated the estimation accuracy of both computers and humans. We found it is possible to equal humans' accuracy using algorithms while also having the potential to reduce biased judgements.
	\item We conducted comprehensive analysis on the relationships between visual features and MI or PI. We answered the research questions about what visual elements an intelligent person will use in profile images, and what visual elements in a profile image make a person perceived to be intelligent. Those insights can guide people to manage their self-representations. 
\end{itemize}


The rest of this paper is organised as follows: Section \ref{sec:related} reviews related works in both psychological and computing literatures. Section \ref{sec:method} introduces the proposed framework including how the MI score is generated and how the MI and PI are estimated by humans and computers. Section \ref{sec:result} presents the experimental results of feature analysis and intelligence estimation. Section \ref{sec:con} draws the conclusions and finally Section \ref{sec:future} summarises the future research directions. 

\section{Related work}
\label{sec:related}
Recent research \cite{kosinski2013private} has examined the extent to which seemingly innocuous social media data such as Facebook status updates \cite{schwartz2013personality} and Facebook `Likes' \cite{youyou2015computer} can be used to infer highly intimate behavioural and psychological traits. Compared to using other types of social media data such as text \cite{schwartz2013personality} for behaviour analysis, the processing of image data is less studied due to the challenge of extracting predictive features (see the survey in \cite{vinciarelli2014a}). Redi \textit{et al.} \cite{redi2015like} predicted the ambiance of commercial places based on the profile pictures of its visitors in Foursquare. Guntuku \textit{et al.} \cite{guntuku2015deep} predicted users' preference for an image based on its visual content and associated tags. Steele \textit{et al.} \cite{steele2009is} explored the human raters' agreement of personality impression from profile images. Liu \textit{et al.} \cite{liu2016analyzing} analysed how Twitter profile images vary with the personality of the users. Cristani \textit{et al.} \cite{cristani2013unveiling} and Segalin \textit{et al.}  \cite{segalin2016the} predicted both measured and perceived personality scores using Flickr images tagged as favourite by users. 

There are very few works studying the relationship between images of people and their intelligence. Kleisner \textit{et al.} \cite{kleisner2014perceived} used geometric morphometrics to determine which facial traits are associated with MI and PI based on 80 facial photographs. Results shown that certain facial traits were shown to be correlated with PI while there was no correlation between morphological traits and MI; PI was associated with MI in men but not women. Talamas \textit{et al.} \cite{talamas2016blinded} investigated PI, attractiveness and academic performance using 100 face photos and found no relationship between PI and academic performance, but a strong positive correlation between attractiveness and PI. They also scrutinized PI and attractiveness in the work in \cite{talamas2016eyelid} based on face photos of 100 adults and 90 school-aged children and found that PI was partially mediated by attractiveness, but independent effects of centre facial cues such as eyelid-openness and subtle smiling can enhance PI scores independent of attractiveness. 

Compared with the above works, we employ social network profile images which include more behavioural cues linking to people's lifestyles, interests, values and thoughts, not just limited to facial cues such as attractiveness or symmetry. We investigate and predict both MI and PI, and also consider demographic information such as gender and age in the analysis. To the best of our knowledge, our work is the first one to automatically estimate intelligence from profile images in social network. 

\section{Proposed framework}
\label{sec:method}


Our framework contains three main elements: 1) users take an IQ test to get their MI scores, 2) users' profile images are shown to other human raters to get PI scores, and 3) visual features are algorithmically extracted from images to estimate MI and PI separately. 

\subsection{Data collection}
\label{subsec:image}


Our data is based on the \textit{myPersonality}\footnote{http://mypersonality.org/} database. \textit{myPersonality} was a Facebook application where users can take psychometric tests by logging in using their Facebook account. Now the database contains more than 6 million psychometric test results together with more than 4 million individual Facebook profiles. About 7,000 users took a 20-item matrices IQ test (validated against Raven's Matrices \cite{raven2003manual}) designed by professional psychometricians; the MI scores of those users are calculated from the IQ test. Their Facebook profiles were collected with consent, which contain demographic information such as gender and age. 

\begin{figure}
	\centerline{\epsfig{figure=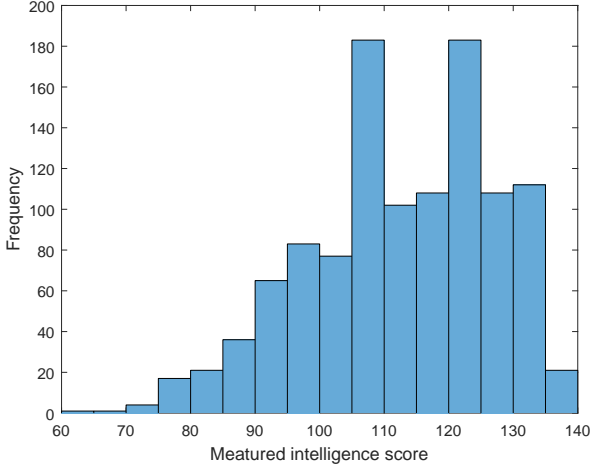,width=0.5\textwidth}}
	\caption{Histogram of MI scores. The collected samples showed a broad range of distribution of MI scores.}
	\label{fig-iqhis}		
\end{figure}

We selected those users who shared their profile images (excluding the Facebook default profile image) and there are 1,122 users (51\% men, age mean$\pm$std=25.9$\pm$9.2, range: 14$\sim$69; MI score mean$\pm$std=112.4$\pm$14.5, range: 64.9$\sim$138.6, distribution is shown in Figure \ref{fig-iqhis}). Each user has one publicly available profile image in JPEG format, normally of size $200\times 200$ pixels. The thumbnails of some example images are shown in Figure \ref{fig-fw}. There are about 16\% non-person images (e.g., cartoons, drawings, animals, signs, etc.), 60\% with only one person, 21\% with two or three persons and 3\% group images (more than four persons). Images with persons may also contain other content such as animals, plants, sports, indoor and outdoor scenes, events (e.g., wedding or graduation), etc.

We also recruited 739 independent raters (49\% men, age mean$\pm$std=24.2$\pm$6.2, range: 15$\sim$72 covering a broad range of raters with different ages) through our online psychometrics platform Concerto\footnote{http://concertoplatform.com/}. Each rater was randomly shown 50 or 100 images, one image at each time. Raters received the following instructions:

\begin{quote}
	At each time, a profile image will be presented to you. The image content may contain (one or more than one) people, animals or other objects. You will be asked to make a judgement of how intelligent this user is who uses such an image as their social network profile image. You need to rate the intelligence score for a given image using a 7-scale point scale wherein 7 stands for the highest ranking (most intelligent) and 1 the lowest (least intelligent). There is no time limit for rating each image.
\end{quote}
    
Each image was finally rated by at least 24 different raters. Each rater's scores were z-score-normalised in order to account for positivity or negativity bias. 

\subsection{Perceived intelligence estimation by humans}
\label{subsec:pibyhuman}

\begin{figure}
	\centerline{\epsfig{figure=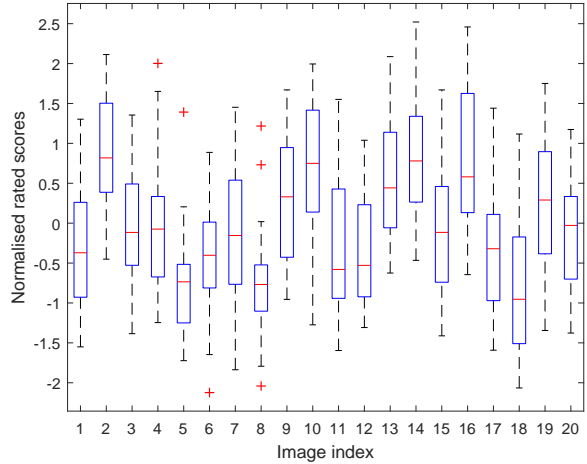,width=0.5\textwidth}}
	\caption{Rated scores for the first 20 images.}
	\label{fig-icc}
\end{figure}

Figure \ref{fig-icc} shows an example of the rated scores for the first 20 images. Each box indicates the distribution of rated scores for each image. The box length is the interquartile range of the scores and the red line indicates the median value. Red crosses indicate some extreme scores where raters strongly disagree with the average. Whiskers indicate the maximal and minimal scores (excluding extreme scores). For each image, there are few extreme scores in the plot and most rated scores tend to cluster in a relatively narrow range. This example shows raters' PI scores are relatively consistent within images but there are differences between images.

In order to quantitatively evaluate whether different raters agree with each other, we compute the inter-rater reliability (IRR) \cite{saal1980rating} for rated scores. The rated scores (observed scores) $X$ is composed of a true score $T$ and a measurement error (noise) $E$ between different raters \cite{mcgraw1996forming}. The IRR analysis aims to determine how much of the variance of the rated scores is due to variance in the true scores after removing the variance of measurement error: 
\begin{equation} 
Reliability = \dfrac{var(T)}{var(T)+var(E)}
\end{equation}
 
To rate all 1,122 images by each rater is time-consuming and raters may lose attention due to long time work. As mentioned in Section \ref{subsec:image}, in our experimental settings, each rater randomly rated 50 or 100 images and each image was finally rated by at least 24 different raters. We use the interclass correction coefficient (ICC) \cite{mcgraw1996forming} to measure the IRR because ICC is suitable when a subset of samples is rated by multiple raters. It models the rated score $X_{ij}$ of image $i$ by rater $j$ as: 
\begin{equation} 
X_{ij} = \mu+r_{i}+e_{ij},
\end{equation} where $\mu$ is the unobserved overall mean which is constant; $r_i$ is an unobserved random effect shared by all rated scores for image $i$ and $e_{ij}$ is an unobserved measurement error. $r_i$ and $e_{ij}$ are assumed to have expected value zero and to be uncorrelated with each other. The variance of $r_j$ is denoted $\sigma_{r}^{2}$ and the variance of $e_{ij}$ is denoted $\sigma_{e}^{2}$.The degree of absolute agreement based on average of $k$ rated scores ($k=24$ in our data) is calculated using the ICC one-way model as (more details see \cite{mcgraw1996forming}):
\begin{equation}
ICC=\dfrac{\sigma_{r}^{2}}{\sigma_{r}^{2}+\frac{\sigma_{e}^{2}}{k}}
\end{equation}

Higher ICC values indicate greater IRR (guideline\cite{cicchetti1994guidelines}: less than 0.4: \textit{poor}, between 0.4 and 0.59: \textit{fair}, between 0.6 and 0.74: \textit{good}, between 0.75 and 1: \textit{excellent}) - in this work it was 0.86. This means that most raters agree with one another in their perception of each image's intelligence. Therefore, the PI score for each image (user) is calculated as the median value of its rated scores. We use median value instead of mean value because the former is more robust to the some extreme rated scores shown in Figure \ref{fig-icc}.

\subsection{Intelligence estimation by computers}

We model the intelligence estimation as a regression problem by mapping image features into MI or PI scores. In this section we first introduce the image features extracted and then present the feature selection schemes used for MI regression and PI regression. After feature selection, the regression is implemented using Support Vector Regression (SVR).    

\subsubsection{Feature extraction}

We use the image aesthetics and perception descriptors \cite{machajdik2010affective,bhattacharya2013towards} inspired by psychology and art theories to extract useful features from profile images. Those features are understandable and effective in human behaviour analysis tasks as shown in previous works \cite{cristani2013unveiling,guntuku2015deep,segalin2016the}. Those features can be classified into four main categories: colour, composition, texture, and body $\&$ face related features. The colour, composition and texture features capture low-level information while the body $\&$ face related features capture the high-level, i.e., content or semantic information. Body is detected using the Deformable Part Model \cite{kokkions2012bounding} and Face related features are extracted by the \textit{Face}$++$ detector \cite{zhou2013extensive}. 

Additionally, as well as extracting the above features from a whole image, we also partition each image into $4\times4$ blocks then extract colour and texture features such as LBP \cite{ahonen2006face} and dense SIFT \cite{lowe2004distinctive}, similar as the work in \cite{guntuku2015do}. LBP is widely used to encode facial texture and dense SIFT is popular for representing scene images. Exacting those features from each block is able to capture the local structure of an image. Section \ref{subsec:fea} will show that there are a large number of such local features significantly correlated with both MI and PI, which can be very useful for intelligence estimation. Details of each feature are shown in Table \ref{tab-fea}.


\begin{table*}[htb]
	\centering
	\begin{threeparttable}
	\caption{\label{tab-fea}Description of image features}
	\begin{tabular}{l|lllp{10cm}}  
		\hlinew{1pt}
		                            & Category                 & Name           & Len. & Description \\
		\hlinew{1pt}
		 \multirow{25}{*}{Low-level}&\multirow{6}{*}{Colour}   & HSV statistics & 5    & Circular variance of H channel, average of S, V (use of light), standard deviation of S, V\\
		                                                       && Emotion-based  & 3    & \textit{Valence}, \textit{Arousal} and \textit{Dominance} in V  and S channels\\
		                                                       && Colourfulness  & 1    & Colour diversity\\
		                                                       && Colour name    & 11   & The percentage of black, blue, brown, grey, orange, pink, purple, red, white and yellow pixels\\                        
		                                                       && Dark channel   & 1    & The minimum filter output on RGB channel, reflects image clarity, saturation and hue\\
		                                                       && Colour sensitivity & 1 & The peak of a weighted colour histogram representing the sensitivity with respect to human eye\\
		 \cline{2-5}
		                            &\multirow{3}{*}{Composition}& Edge pixels     & 1 & The percentage of edge pixels to present the structure of an image\\
		                                                         && Regions         & 2 & Number of regions, average size of regions\\
		                                                         && Symmetry        & 2 & Horizontal symmetry and vertical symmetry\\
		 \cline{2-5}
		                            &\multirow{6}{*}{Texture}& Entropy  & 1 & Gray distribution entropy\\
		                                                     && Sharpness& 4 & The average, variance, minimal and maximal value of sharpness\\
		                                                     && Wavelet  & 12 &  Wavelet textures (spatial smoothness/graininess) in 3 levels on each HSV channel\\
                                                             && Tamura   & 3 & Coarseness, contrast and directionality of texture\\ 
		                                                     && GLCM  & 12 & Contrast, correlation, energy, homogeneousness for each HSV channel\\
		                                                     && GIST  & 24 & Low dimensional representation of a scene, extracting from a whole image\\
		 \cline{2-5}
		                           &\multirow{4}{*}{Local} & Colour histogram & 512 & Histogram of colour from local blocks \\
		                                                   && LBP              & 944 & Local Binary Pattern ($LBP_{i,2}^{u2}$) from local blocks \\
		                                                   && GIST             & 512 & GIST features extracted from local blocks \\
		                                                   && SIFT             & 2048 & Dense SIFT features from local blocks \\    
		  \hline
	      \multirow{5}{*}{High-level}&\multirow{5}{*}{Body \& face} & Body & 2 & The presence of body* and the proportion of the main body \\
	                                                                && Skin   & 1 & The percentage of skin pixels \\
	                                                                && Face   & 4 & The number of faces*, the proportion of main face, the horizontal and vertical locations of main face \\   
	                                                                && Glasses& 2 & The presence of normal glasses* or sunglasses* \\
	                                                                && Pose   & 3 & The pitch angle, roll angle and yaw angle of head \\ 
		\hlinew{1pt}
	\end{tabular}
			\begin{tablenotes}
				\footnotesize
				\item $*$ with manual check to make sure the automatic detection results are correct
			\end{tablenotes}
		\end{threeparttable}
\end{table*}
	
\subsubsection{Feature selection}
\label{subsubsec:fs}

After feature extraction, all features of an image are concatenated into an M-dimensional vector. We employ Principle Component Analysis (PCA) to reduce feature dimension by selecting $N$ dimensions which have the highest amount of variance where $N$ is the number of training samples ($N\ll M$). 

Next, we adopt filter based feature selection on the uncorrelated PCA features to select the most informative features. In psychometrics tests, MI scores can be classified into several categories according to the values. For example:  
\begin{quote}
1-very superior: $\geq130$\\
2-superior: 120$\sim$129\\
3-high average: 110$\sim$119\\
4-average: 90$\sim$109\\
5-low average: 80$\sim$89\\
6-borderline: 70$\sim$79\\
7-low: $\leq69$ 
\end{quote}

Inspired by this, we choose the most discriminant features while constraining the same feature to be selected across different categories of MI scores. This is similar to the idea of multi-task learning \cite{zhu2014matrix} which learns models for multiple tasks jointly using a shared representation. 

We perform univariate statistical tests on features and the target variable in the training set and select the most statistically significant $K$ features (according to p-value/F-score by F-test). For MI, we select $K$ features based on MI score and another $K$ features based on MI label and then use the intersection of those two feature sets as the final selected features. For PI, the scores are already calculated based on a 7-point scale, so the target variable is the PI score directly.




\section{Experiments and results}
\label{sec:result}
Parameters for image feature extraction are set as the optimal values reported in their original papers. We present univariate correlations between features and intelligence scores and then report the overall intelligence estimation results. Experiments are performed using the Leave-one-out cross-validation: Each time, one image is selected as testing sample and the remaining images are training samples on which regression models are trained. The selected dimension $K$ in section \ref{subsubsec:fs} is set as $N/2$ where $N$ is the number of training samples. We use cross-validation in training set to find the optimal parameters for SVR to avoid overfitting.     
\newline
\subsection{Feature analysis}
\label{subsec:fea}

We calculate the Spearman correlation coefficient $\rho$ between each individual feature and MI or PI scores. In this section we describe: 1) what visual elements an intelligent person will use in profile images, and 2) what visual elements in a profile image make a person perceived to be intelligent.    

As shown in Table \ref{tab-ratio}, the ratios of significantly (p value <0.05) correlated features with PI are much larger than that with MI. This might be because PI scores are based on only raters' perceptions of a user's profile image, but MI scores may involve other factors which cannot be directly inferred from images such as education level. Most of the colour, composition, body and face, and texture features are significantly correlated with MI and PI. Although the ratios of local features are relatively low, considering the high dimension of those features there are still a large number of features which are useful for regression. 

Correlations between individual features and MI or PI are shown in Figure \ref{fig-fea}. For local features, since the dimensions are very high, only features with significant correlations are selected in the calculation and the average positive and negative correlations are shown. For face and body features, only images with faces are included.

\subsubsection{Colour}
The percentage of green pixels and colour sensitivity are positively correlated with MI while H circular variance (colour diversity in hue), percentages of pink, purple and red pixels, and dark channel (the higher the value, the less clear the image) are negatively correlated with MI. This suggests people with high MI scores like to represent themselves in a clear and simple profile image. Percentages of grey and white pixels are positively correlated with PI, which suggests those colours in a profile image makes the user look intelligent. H circular variance, average of S channel (indicates chromatic purity, the lower the value, the purer the colour), percentages of brown, green, pink, purple pixels and dark channel are negatively correlated with PI.   

\subsubsection{Composition}
Percentage of edge pixels is negatively correlated with PI. Most other composition features are correlated with neither MI nor PI. This suggests that in this dataset people do not prefer particular composition elements in their profile images. Moreover, those elements also do not affect how others judge the user's intelligence.

\subsubsection{Body \& face}
The proportions of skin and face area are negatively correlated with MI. Similarly, the proportions of body, skin and face are negatively correlated with PI. Users who are very close to the camera are perceived as less intelligent. The smile degree and the presence of glasses are positively correlated with PI while no such correlations are observed in MI, indicating that people whose pictures show them smiling and wearing glasses are perceived as being intelligent but this is an inaccurate stereotype.

\subsubsection{Texture}
The average, minimal and maximal values of sharpness are positively correlated with both MI and PI. The variance of sharpness is negatively correlated with PI. This confirms that the sharper (more clear in texture) a profile image is, the more intelligent the user is perceived to be. Most wavelet features, which represent graininess, are negatively correlated with MI. Most GIST features, which represent the dominant spatial structure of a scene, are positively correlated with both MI and PI.  

\subsubsection{Local}
The average positive and negative correlations are shown separately. The texture features LBP achieve the highest correlations for both MI and PI.

\begin{table}[htb]
	\scriptsize
	\centering
	\caption{\label{tab-ratio}Ratios of significantly correlated feature ($p<0.05$)}
	\begin{tabular}{llll}  
		\hlinew{1pt}
		\multirow{2}{*}{Feature}                 & \multirow{2}{*}{Len.}&\multicolumn{2}{l}{$p<0.05$ ratio} \\
		\cline{3-4}
	    & &  MI    & PI \\
		\hlinew{1pt}
	Colour, composition, body\&face, texture & 95 & 34.3\% & 56.3\% \\
	Local colour histogram                   & 512 & 8.0\%  & 11.9\% \\
	Local LBP                                & 944 & 15.4\% & 44.6\% \\
	Local GIST                               & 512 & 9.0\%  & 43.2\% \\
	Local SIFT                               & 2048& 7.4\%  & 11.7\%\\ 
		\hlinew{1pt}
	\end{tabular}
\end{table}

\begin{figure*}[htb]
	\centerline{\epsfig{figure=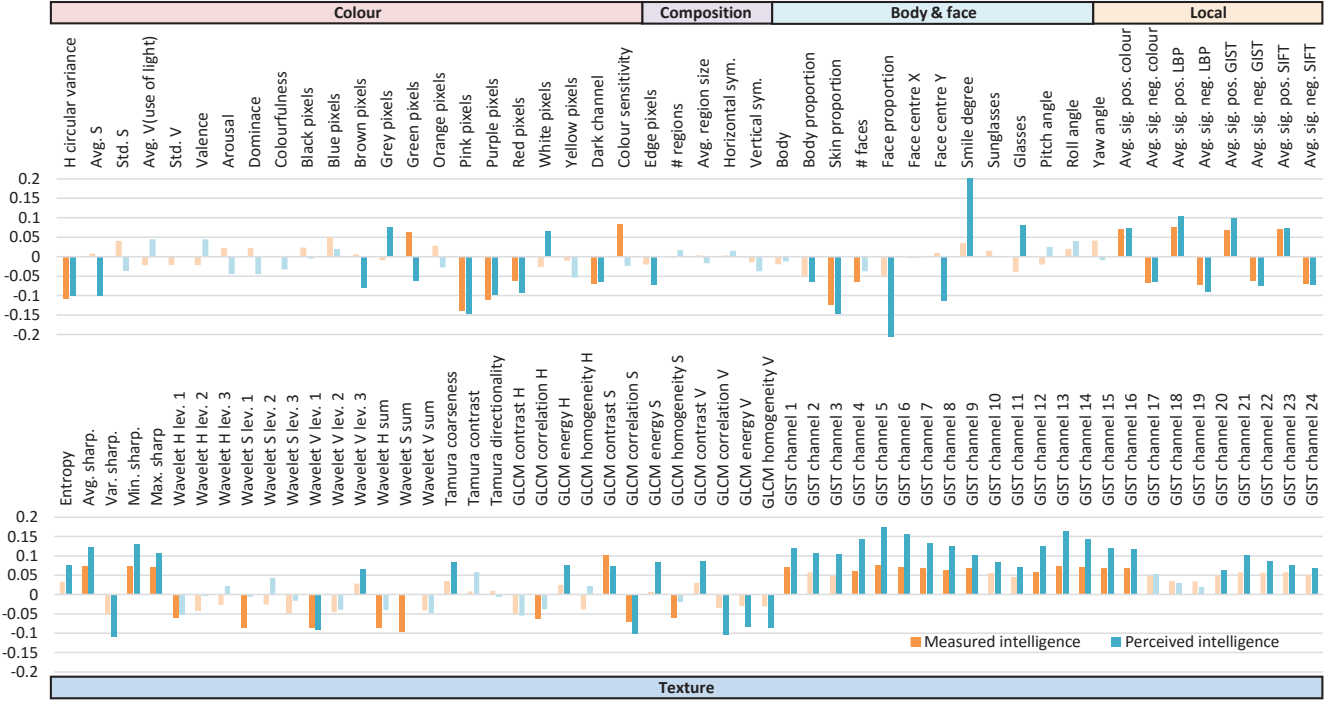,width=1\textwidth}}
	\caption{Correlations between image features and MI (orange bars) or PI (blue bars). Darker bars indicate correlations which are statistically significant ($p<0.05$).}
	\label{fig-fea}
\end{figure*}

\subsection{Intelligence estimation}
\subsubsection{By humans}

The previous research \cite{talamas2016blinded} which is based on 100 facial photographs showed that PI is associated MI in men (photo) but not women (photo). Here we also partition users into two groups by gender and analyse whether PI can predict MI, i.e., to see is there any difference when human raters make intelligence judgement for male users and female users. 


For each user group (i.e., male user or female user), we run the linear regression to test the association between MI (dependent variable) and PI (independent variable) while taking the age of users as controlled variable. We found that PI significantly affects MI in both male and female users (both p<0.001). For better comparison, we report the correlation between PI and MI for male and female users separately as shown in Table \ref{tab-human}. PI are significantly correlated with MI for both male and female users. This is because the work in \cite{talamas2016blinded} uses images containing only cropped faces and raters are more likely affected by attractiveness of the face. Our profile images contains more behavioural cues, as mentioned in Section \ref{sec:related}, which provides more information for raters to make their intelligence judgements.  But we also confirm the observation that the correlation for female users are lower than that for male users in all rater groups (i.e., male, female and together).


\begin{table}[htb]
	\centering
	\begin{threeparttable}
		\caption{\label{tab-human}Spearman correlation $\rho$ between PI and MI}
		\begin{tabular}{llll}  
			\hlinew{1pt}        
			               & Male users   & Female users & Together \\
			\hlinew{1pt}
			Male raters    & 0.23 & 0.21 & 0.24  \\
			Female raters  & 0.23& 0.18 & 0.22  \\
		        Together       & 0.25& 0.20 & 0.24  \\
			\hlinew{1pt}
		\end{tabular}
		\begin{tablenotes}
			\footnotesize
			\item All correlations are significant at p <0.001 level
		\end{tablenotes}
	\end{threeparttable}
\end{table}

\subsubsection{By computers}
We report both the MI and PI estimation results in terms of correlation and root-mean-square error (RMSE) as shown in Table \ref{tab-results}. Since MI and PI scores are in different scales, we also report the Normalized RMSE to allow comparison (i.e., $RMSE/(\boldsymbol{y}_{max}-\boldsymbol{y}_{min})$ where $\boldsymbol{y}$ is the actual MI or PI scores). Similar to previous work \cite{segalin2016the}, we compare the performance of human and computer with two methods which generate predicted scores from: 1) Random: random scores from a Gaussian distribution with the same mean and variance of training scores (we run this ten times and reported the average results), and 2) Mean: the average of training scores, which can minimise the RMSE.


\begin{table}[htb]
	\centering
	\begin{threeparttable}
		\caption{\label{tab-results}Estimation results}
		\begin{tabular}{lllll}  
			\hlinew{1pt}        
			                  & Spearman $\rho$ & RMSE & NRMSE \\
			\hlinew{1pt}
			MI               &              & \\
			\hline 
			Human       & $0.24^{***}$      & --    & -- \\
			Computer  & $0.27^{***}$      & 14.50 & 0.20\\ 
			Random        & $<0^{*}$      & 15.13& 0.21\\
			Mean          & --                  & 14.49 & 0.20\\ 
			\hline
			PI  &              & \\
			\hline
			Computer                   & $0.36^{***}$    & 0.54  & 0.15\\ 
			Random        & $<0^{**}$         & 0.58 & 0.17\\
			Mean          & --       & 0.56   & 0.16\\ 
			\hlinew{1pt}
		\end{tabular}
		\begin{tablenotes}
			\scriptsize
			\item $***: p<0.001, **: p<0.01, *: p<0.05$
		\end{tablenotes}
	\end{threeparttable}
\end{table}

For MI, the computer's estimation ($\rho=0.27$) is slightly better than that of humans ($\rho=0.24$). Intelligence estimation from images is a difficult task even for humans, but it is possible to use algorithms to estimate it beyond a random guess. For PI, the correlation is higher ($\rho=0.36$), since to some extent, the computer is able to extract effective visual features may be used by humans as shown in Figure \ref{fig-fea}. 

The RMSE of computer is lower than that of the Random, but not (much) lower than the Mean. The performance improvement of computer is much better in terms of correlation than in terms of RMSE is because the regressor tends to predict the \textit{rank} between different scores better in terms of Spearman's coefficient other than the actual \textit{values}. Similar observation has been shown in \cite{segalin2016the} and this is also compatible with the prevailing opinion on intelligence in psychology that IQ is essentially a rank other than a true unit of intellectual ability \cite{bartholomew2004measuring}. 

\subsection{Discussion}
\subsubsection{Human bias}
Given that the computer's predictions of a user's intelligence do not match humans' perceptions, it begs the question of what features humans use that our image features do not capture. For example, the more complicated high-level semantic features.   
We manually examined the 50 images with the highest and lowest PI scores (due to the privacy agreement with participants, we are unable to show example images of those two groups). Inevitably this is a subjective exercise but we hope that the following insights will be helpful to guide future algorithms. 

The top 50 images which are perceived as high intelligence contain visual cues related to business clothing, books, instruments, chess, science, music, art, university, formal dining, archery, computers, and math. The bottom 50 images contain visual cues such as colourful hair, offensive hand gestures, an overweight body, heavy make-up, smokers, tattoos, and black ethnicity. 

Together with the results in Section \ref{subsec:fea}, which found that humans use cues, such as wearing glasses, that are uncorrelated with MI, this provides further evidence of how human raters tend to use biased visual cues when judging intelligence from profile images. In contrast, computing approaches may have the potential to reduce human bias.

\subsubsection{Application}
Profile pictures are a ubiquitous way for users to present themselves on social networks, and typically they are not considered to be private data. Based on our work, an automatic profile picture rating system could be created to determine whether a profile image ``looks'' intelligent and suitable for professional places such as LinkedIn and on a CV. This can help people to better manage their self-representations. Commercial companies can similarly benefit from such an intelligence rating system by selecting suitable portrait photos in brand representation to make a good impression on customers. 


\section{Conclusion}
\label{sec:con}

We have proposed a framework of intelligence estimation from profile images both by humans and computers. Humans are able to make intelligence judgements for both men and women from their social network profile images. Experimental results show that it is possible to equal humans' accuracy using computing methods, while also having the potential to reduce biased judgements based on inaccurate stereotypes. 

Our results found that people who are measured as intelligent and also perceived as intelligent do not like to use the colour pink, purple or red in their profile images, and their images are usually less diversified in colour, more clear in texture, and contain less skin area. Besides that, intelligent people also like to use the colour green, and have fewer faces in their images, but this does not affect how others judge them. Essentially, the most intelligent people in our dataset understand that a profile picture is most effective when it shows a single person, captured in focus, and with an uncluttered background.

The following cues are inaccurate stereotypes - correlated with perceptions of intelligence but not measured intelligence. Profile images containing more grey and white, but less brown and green, with higher chromatic purity, smiling and wearing glasses, and faces at a proper distance from the camera, make people look intelligent no matter how smart they really are.

Our results show that humans use inaccurate stereotypes to make biased judgements about the perceived intelligence of the person who uses a profile picture, which raises considerable concerns about the common practice for hiring managers to search candidates online before inviting them to interview. Nevertheless, our results also indicate that the choices that users make of how to present themselves in their profile picture are reflective of their measured intelligence, and that computer algorithms can extract features and automatically make predictions.

\section{Future work}
\label{sec:future}
There are several directions for future work. 1) Our results show that high-level features which capture the content and semantic information of images are helpful for intelligence estimation. In the future we will investigate additional high-level features with more sophisticated object detection and image caption techniques. Those interpretable features will benefit the human behaviour research in social science.  2) Psychology research showed that individual differences are influenced by both biological and environmental factors \cite{triandis2002cultural}. So it also should consider sociodemographic factors such as ethnicity, language and culture in the analysis for visual features and this will help generate prior knowledge for better feature selection. 3) Other efforts can focus on improving the estimation accuracy such as collecting multiple images for each user then modelling the estimation process using multiple instance regression models or fusing results by multiple classifier system in different levels (e.g., feature, score, rank, etc.). 
%

%
	
\bibliographystyle{abbrv}
\bibliography{IEEEabrv,sigproc} 

\begin{thebibliography}{10}

\bibitem{ahonen2006face}
T.~Ahonen, A.~Hadid, and M.~Pietikainen.
\newblock Face description with local binary patterns: Application to face
  recognition.
\newblock {\em {IEEE} Trans. Pattern Anal. Mach. Intell.}, 28(12):2037--2041,
  Dec 2006.

\bibitem{baert2015do}
S.~Baert.
\newblock Do they find you on facebook? facebook profile picture and hiring
  chances.
\newblock {\em IZA Discussion Paper}, 9584, 2015.

\bibitem{bartholomew2004measuring}
D.~J. Bartholomew.
\newblock {\em Measuring intelligence: Facts and fallacies}.
\newblock Cambridge University Press, 2004.

\bibitem{bhattacharya2013towards}
S.~Bhattacharya, B.~Nojavanasghari, T.~Chen, D.~Liu, S.-F. Chang, and M.~Shah.
\newblock Towards a comprehensive computational model foraesthetic assessment
  of videos.
\newblock In {\em {ACM} Int. Conf. Multimedia ({MM})}, pages 361--364, 2013.

\bibitem{cicchetti1994guidelines}
D.~V. Cicchetti.
\newblock Guidelines, criteria, and rules of thumb for evaluating normed and
  standardized assessment instruments in psychology.
\newblock {\em Psychol. assessment}, 6(4):284, 1994.

\bibitem{cristani2013unveiling}
M.~Cristani, A.~Vinciarelli, C.~Segalin, and A.~Perina.
\newblock Unveiling the multimedia unconscious: Implicit cognitive processes
  and multimedia content analysis.
\newblock In {\em {ACM} Int. Conf. Multimedia ({MM})}, pages 213--222, 2013.

\bibitem{guntuku2015do}
S.~C. Guntuku, L.~Qiu, S.~Roy, W.~Lin, and V.~Jakhetiya.
\newblock Do others perceive you as you want them to?: Modeling personality
  based on selfies.
\newblock In {\em Int. Workshop on Affect and Sentiment in Multimedia {ASM}},
  pages 21--26, 2015.

\bibitem{guntuku2015deep}
S.~C. Guntuku, J.~T. Zhou, S.~Roy, L.~Weisi, and I.~W. Tsang.
\newblock {\em Asian Conf. Computer Vision ({ACCV})}, chapter Deep
  Representations to Model User `Likes', pages 3--18.
\newblock 2015.

\bibitem{keller2013the}
M.~C. Keller, C.~E. Garver-Apgar, M.~J. Wright, N.~G. Martin, R.~P. Corley,
  M.~C. Stallings, J.~K. Hewitt, and B.~P. Zietsch.
\newblock The genetic correlation between height and {IQ}: Shared genes or
  assortative mating?
\newblock {\em PLoS Genet}, 9(4):1--10, 04 2013.

\bibitem{kleisner2014perceived}
K.~Kleisner, V.~Chv$\acute{a}$talov$\acute{a}$, and J.~Flegr.
\newblock Perceived intelligence is associated with measured intelligence in
  men but not women.
\newblock {\em PLoS ONE}, 9(3):1--7, 03 2014.

\bibitem{kokkions2012bounding}
I.~Kokkinos.
\newblock Bounding part scores for rapid detection with deformable part models.
\newblock In {\em European Conf. Computer Vision ({ECCV})}, pages 41--50, 2012.

\bibitem{kosinski2013private}
M.~Kosinski, D.~Stillwell, and T.~Graepel.
\newblock Private traits and attributes are predictable from digital records of
  human behavior.
\newblock {\em {PNAS}}, 110(15):5802--5805, 2013.

\bibitem{liu2016analyzing}
L.~Leqi, D.~Preo{\c t}iuc-Pietro, Z.~Riahi, M.~E. Moghaddam, and L.~Ungar.
\newblock Analyzing personality through social media profile picture choice.
\newblock In {\em Int. {AAAI} Conf. Web and Social Media ({ICWSM})}, 2016.

\bibitem{lowe2004distinctive}
D.~G. Lowe.
\newblock Distinctive image features from scale-invariant keypoints.
\newblock {\em Int. J. Comput. Vision}, 60(2):91--110, 2004.

\bibitem{machajdik2010affective}
J.~Machajdik and A.~Hanbury.
\newblock Affective image classification using features inspired by psychology
  and art theory.
\newblock In {\em {ACM} Int. Conf. Multimedia ({MM})}, pages 83--92, New York,
  NY, USA, 2010. ACM.

\bibitem{mcgraw1996forming}
K.~O. McGraw and S.~P. Wong.
\newblock Forming inferences about some intraclass correlation coefficients.
\newblock {\em Psychol. methods}, 1(1):30, 1996.

\bibitem{raven2003manual}
J.~Raven, J.~Raven, and J.~Court.
\newblock Manual for raven's progressive matrices and vocabulary scales.
\newblock {\em San Antonio, TX: Harcourt Assessment}, 2003.

\bibitem{redi2015like}
M.~Redi, D.~Quercia, L.~Graham, and S.~Gosling.
\newblock Like partying? your face says it all. predicting the ambiance of
  places with profile pictures.
\newblock In {\em Int. {AAAI} Conf. Web and Social Media ({ICWSM})}, 2015.

\bibitem{saal1980rating}
F.~E. Saal, R.~G. Downey, and M.~A. Lahey.
\newblock Rating the ratings: Assessing the psychometric quality of rating
  data.
\newblock {\em Psychol. Bull.}, 88(2):413, 1980.

\bibitem{schwartz2013personality}
H.~A. Schwartz, J.~C. Eichstaedt, M.~L. Kern, L.~Dziurzynski, S.~M. Ramones,
  M.~Agrawal, A.~Shah, M.~Kosinski, D.~Stillwell, M.~E.~P. Seligman, and L.~H.
  Ungar.
\newblock Personality, gender, and age in the language of social media: The
  open-vocabulary approach.
\newblock {\em PLoS ONE}, 8(9):1--16, 09 2013.

\bibitem{segalin2016the}
C.~Segalin, A.~Perina, M.~Cristani, and A.~Vinciarelli.
\newblock The pictures we like are our image: Continuous mapping of favorite
  pictures into self-assessed and attributed personality traits.
\newblock {\em {IEEE} Trans. Affect. Comput.}, PP(99):1--1, 2016.

\bibitem{steele2009is}
F.~Steele, D.~C. Evans, and R.~K. Green.
\newblock Is your profile picture worth 1000 words? photo characteristics
  associated with personality impression agreement.
\newblock In {\em Int. {AAAI} Conf. Web and Social Media ({ICWSM})}, 2009.

\bibitem{talamas2016eyelid}
S.~N. Talamas, K.~I. Mavor, J.~Axelsson, T.~Sundelin, and D.~I. Perrett.
\newblock Eyelid-openness and mouth curvature influence perceived intelligence
  beyond attractiveness.
\newblock {\em J. Exp. Psychol.-Gen.}, 145(5):603, 2016.

\bibitem{talamas2016blinded}
S.~N. Talamas, K.~I. Mavor, and D.~I. Perrett.
\newblock Blinded by beauty: Attractiveness bias and accurate perceptions of
  academic performance.
\newblock {\em PLoS ONE}, 11(2):1--18, 02 2016.

\bibitem{triandis2002cultural}
H.~C. Triandis and E.~M. Suh.
\newblock Cultural influences on personality.
\newblock {\em Annu. rev. psychol.}, 53(1):133--160, 2002.

\bibitem{vinciarelli2014a}
A.~Vinciarelli and G.~Mohammadi.
\newblock A survey of personality computing.
\newblock {\em {IEEE} Trans. Affect. Comput.}, 5(3):273--291, July 2014.

\bibitem{youyou2015computer}
W.~Youyou, M.~Kosinski, and D.~Stillwell.
\newblock Computer-based personality judgments are more accurate than those
  made by humans.
\newblock {\em {PNAS}}, 112(4):1036--1040, 2015.

\bibitem{zagorsky2007do}
J.~L. Zagorsky.
\newblock Do you have to be smart to be rich? the impact of {IQ} on wealth,
  income and financial distress.
\newblock {\em Intelligence}, 35(5):489 -- 501, 2007.

\bibitem{zhou2013extensive}
E.~Zhou, H.~Fan, Z.~Cao, Y.~Jiang, and Q.~Yin.
\newblock Extensive facial landmark localization with coarse-to-fine
  convolutional network cascade.
\newblock In {\em {IEEE} Int. Conf. Computer Vision Workshops ({ICCVW})}, pages
  386--391, Dec. 2013.

\bibitem{zhu2014matrix}
X.~Zhu, H.~I. Suk, and D.~Shen.
\newblock Matrix-similarity based loss function and feature selection for
  alzheimer's disease diagnosis.
\newblock In {\em {IEEE} Conf. Computer Vision and Pattern Recognition
  ({CVPR})}, pages 3089--3096, June 2014.

\end{thebibliography}

%
%

\end{document}